\pdfoutput=1

\documentclass[11pt]{article}

\usepackage[]{ACL2023}

\usepackage{times}
\usepackage{latexsym}

\usepackage[T1]{fontenc}

\usepackage[utf8]{inputenc}

\usepackage{microtype}
\usepackage{amsmath}
\usepackage{amsfonts}
\usepackage{amssymb}
\usepackage{pifont}
\usepackage{subfig}
\usepackage{graphicx}
\usepackage{booktabs}
\usepackage{multirow}
\usepackage{bm}
\usepackage{csquotes}
\usepackage{xcolor,colortbl}
\usepackage{afterpage}

\DeclareMathOperator*{\argmax}{arg\,max}

\definecolor{applegreen}{rgb}{0.55, 0.71, 0.0}
\definecolor{bleudefrance}{rgb}{0.19, 0.55, 0.91}
\definecolor{darkmagenta}{rgb}{0.55, 0.0, 0.55}
\definecolor{davysgrey}{rgb}{0.33, 0.33, 0.33}
\definecolor{brickred}{rgb}{0.8, 0.25, 0.33}
\definecolor{brightred}{HTML}{E06666}
\newif\ifcomments
\commentstrue
\ifcomments
    \providecommand\liat[1]{[\textcolor{brown}{Liat: {#1}}]}
    \providecommand\adi[1]{[\textcolor{purple}{Adi: {#1}}]}
    \providecommand\jb[1]{[\textcolor{blue}{Jonathan: {#1}}]}
    \providecommand\amir[1]{[\textcolor{orange}{Amir: {#1}}]}
    \providecommand\ori[1]{[\textcolor{red}{Ori: {#1}}]}
    \providecommand\yb[1]{[\textcolor{applegreen}{Yonatan: {#1}}]}
\else
    \providecommand{\liat}[1]{}
    \providecommand{\adi}[1]{}
    \providecommand{\jb}[1]{}
    \providecommand{\amir}[1]{}
    \providecommand{\ori}[1]{}
    \providecommand{\yb}[1]{}
\fi

\title{
\emph{What Are You Token About?}\\ Dense Retrieval as Distributions Over the Vocabulary}

\author{Ori Ram$^1$~~~~Liat Bezalel$^1$~~~~Adi Zicher$^1$\\\textbf{Yonatan Belinkov$^2$\thanks{~~Supported by the Viterbi Fellowship in the Center for Computer Engineering at the Technion.}~~~~Jonathan Berant$^1$~~~~Amir Globerson$^1$}\\ 
$^1$Blavatnik School of Computer Science, Tel Aviv University~~~$^2$Technion -- IIT, Israel\\
\small{\texttt{ori.ram@cs.tau.ac.il, liatbezalel@mail.tau.ac.il, adizicher@mail.tau.ac.il}}\\
\small{\texttt{ belinkov@technion.ac.il, joberant@cs.tau.ac.il, gamir@tauex.tau.ac.il}}
}

\begin{document}
\maketitle

\begin{abstract}
Dual encoders are now the dominant architecture for dense retrieval. Yet, we have little
understanding of how they represent text, and why this leads to good performance.
In this work, we shed light on this question via \textit{distributions over the vocabulary}.
We propose to interpret the vector representations produced by dual encoders by projecting them into the model's vocabulary space. 
We show that the resulting projections contain rich semantic information, and draw connection between them and sparse retrieval.
We find that this view can offer an explanation for some of the failure cases of dense retrievers. For example, we observe that the inability of models to handle tail entities is correlated with a tendency of the token distributions to \textit{forget} some of the tokens of those entities. We leverage this insight and propose a simple way to \textit{enrich} query and passage representations with lexical information at \emph{inference} time, and show that this significantly improves performance compared to the original model in zero-shot settings, and specifically on the BEIR benchmark.\footnote{Our code is publicly available at 
\url{https://github.com/oriram/dense-retrieval-projections}.
}
\end{abstract}
\begin{figure}[t]
\centering
\hspace*{-20pt}
\includegraphics[width=1.05\columnwidth]{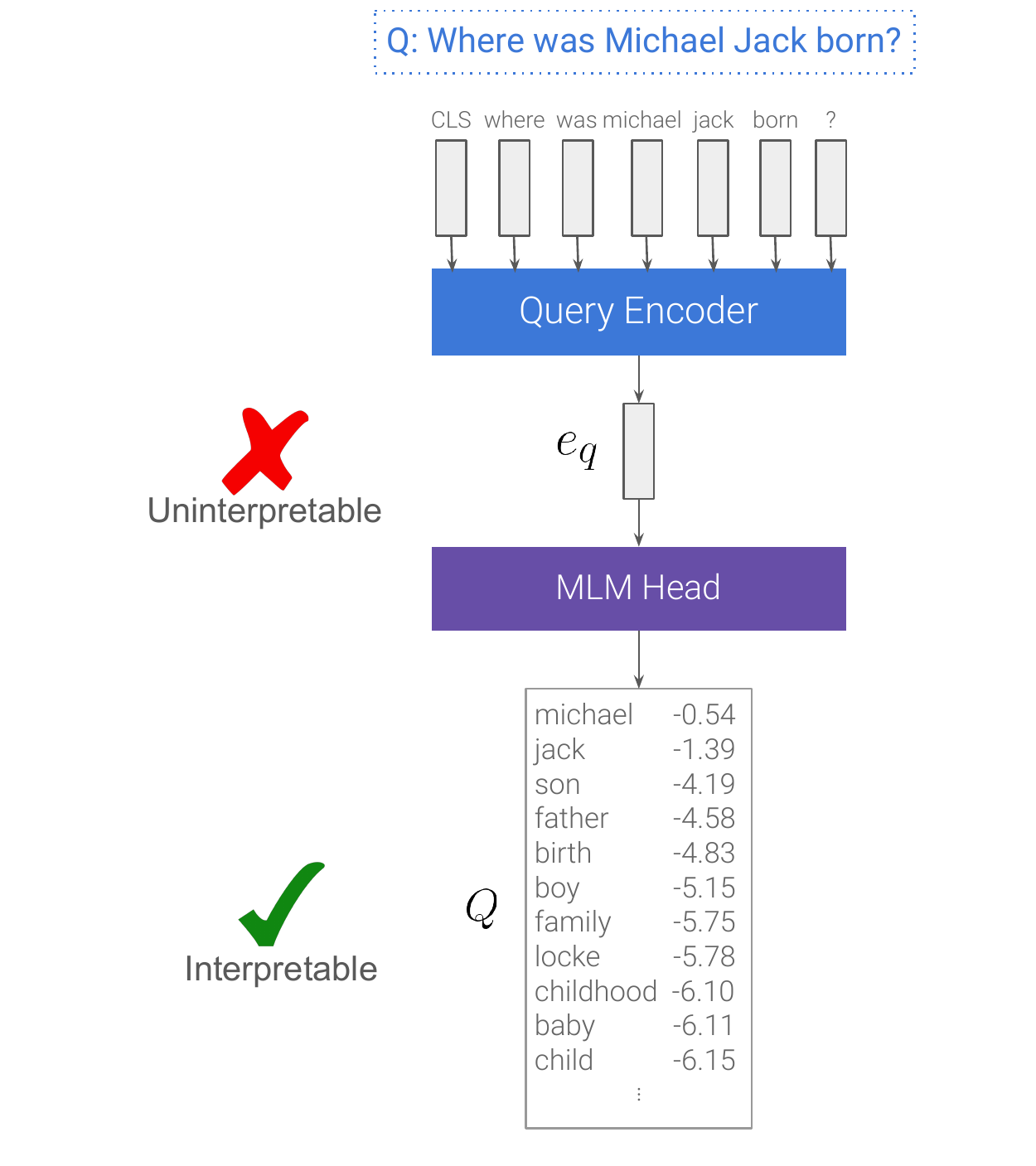}
\vspace{-5pt}
\caption{
An example of our framework. 
We run the question ``\textit{Where was Michael Jack born?}'' through the question encoder of DPR \cite{karpukhin-etal-2020-dense}, and project the question representation $\bm{e}_q$ to the vocabulary space using BERT's masked language modeling head \cite{devlin-etal-2019-bert}. The result is a distribution over the vocabulary, $Q$. We apply the same procedure for passages as well. 
These projections enable reasoning about and improving retrieval representations. 
}
\label{fig:example_intro}
\end{figure}
\begin{figure*}[t!]
\vspace{-10pt}
\centering
\includegraphics[width=\textwidth]{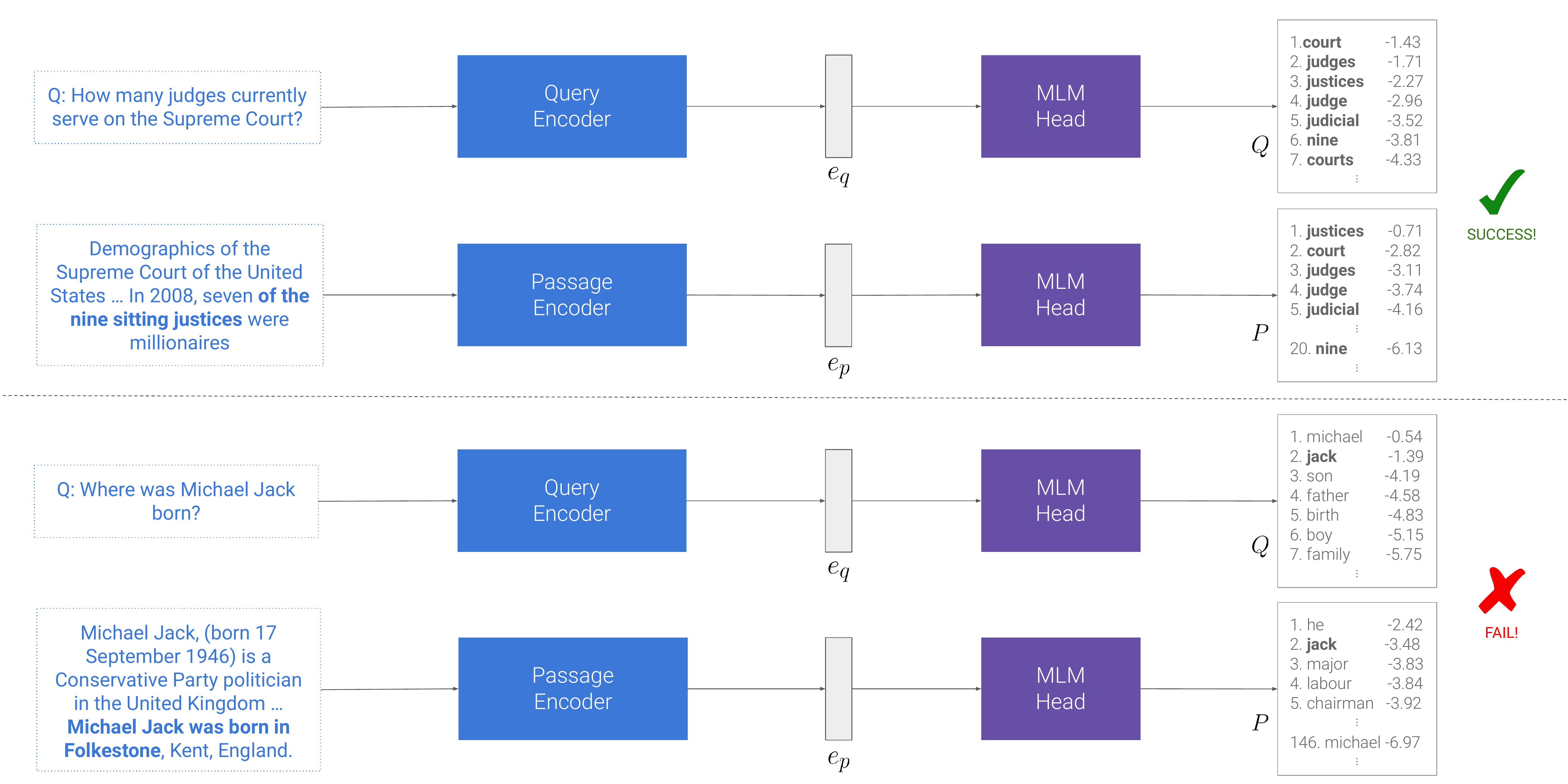}
\caption{A success case from Natural Questions (top) and a failure case from EntityQuestions (bottom) of DPR \cite{karpukhin-etal-2020-dense}, explained via projecting question and (its relevant) passage representations to the vocabulary space.
Tokens in the top-20 of both question and passage vocabulary projections are marked in bold.
}
\label{fig:intro_two_examples}
\end{figure*}

\section{Introduction}
Dense retrieval models based on neural text representations have proven very effective \cite{karpukhin-etal-2020-dense,qu-etal-2021-rocketqa,ram-etal-2022-learning,izacard2022unsupervised,izacard2022atlas}, improving upon strong traditional sparse models like BM25 \cite{bm25}. 
However, when applied off-the-shelf (\textit{i.e.}, in \textit{out-of-domain} settings) they often experience a severe drop in performance \cite{beir,sciavolino-etal-2021-simple,reddy2021robust}.
Moreover, the reasons for such failures are poorly understood, as the information captured in their representations remains under-investigated. 

In this work, we present a new approach for interpreting and reasoning about dense retrievers, through distributions induced by their query\footnote{Throughout the paper, we use \textit{query} and \textit{question} interchangeably.} and passage representations when projected to the vocabulary space, namely distributions over their vocabulary space (Figure~\ref{fig:example_intro}). 
Such distributions enable a better understanding of the representational nature of dense models and their failures, which paves the way to simple solutions that improve their performance. 

We begin by showing that dense retrieval representations can be projected to the vocabulary space, by feeding them through the masked language modeling (MLM) head of the pretrained model they were initialized from \textit{without any further training}. This operation results in distributions over the vocabulary, which we refer to as \textit{query vocabulary projections} and \textit{passage vocabulary projections}.

Surprisingly, we find these projections to be  highly interpretable to humans (Figure~\ref{fig:intro_two_examples}; Table~\ref{tab:examples}).
We analyze these projections and draw interesting connections between them and well-known concepts from sparse retrieval (\S\ref{sec:analysis}).
First, we highlight the high coverage of tokens shared by the query and the passage in the top-$k$ of their projections. This obersvation suggests that the \textit{lexical overlap} between query and passages plays an important role in the retrieval mechanism.
Second, we show that vocabulary projections of passages they are likely to contain words that appear in queries about the given passage. Thus, they can be viewed as predicting the questions one would ask about the passage.
Last, we show that the model implicitly implements \textit{query expansion} \cite{relevance-feedback}. For example, in Figure~\ref{fig:intro_two_examples} the query is ``\textit{How many judges currently serve on the Supreme court?}'', and the words in the query projection $Q$ include ``\textit{justices}'' (the common way to refer to them) and  ``\textit{nine}'' (the correct answer).

The above findings are especially surprising due to the fact that these retrieval models are fine-tuned in a contrastive fashion, and thus do not perform any prediction over the vocabulary or make any use of their language modeling head during fine-tuning. In addition, these representations are the result of running a deep transformer network that can implement highly complex functions. Nonetheless, model outputs remain ``faithful'' to the original lexical space learned during pretraining.

We further show that our approach is able to shed light on the reasons for which dense retrievers struggle with simple entity-centric questions \cite{sciavolino-etal-2021-simple}. Through the lens of vocabulary projections, we identify an interesting phenomenon: dense retrievers  tend to ``ignore'' some of the tokens appearing in a given passage. This is reflected in the ranking assigned to such tokens in the passage projection. For example, the word ``\textit{michael}'' in the bottom example of Figure~\ref{fig:intro_two_examples} is ranked relatively low (even though it appears in the passage title), thereby hindering the model from retrieving this passage.  
We refer to this syndrome as \textit{token amnesia} (\S\ref{sec:token_forgetfulness}).

We leverage this insight and suggest a simple inference-time fix that enriches dense representations with lexical information, addressing token amnesia.
We show that lexical enrichment significantly improves performance compared to vanilla models on the challenging BEIR benchmark \cite{beir} and additional datasets. 
For example, we boost the performance of the strong MPNet model on BEIR from 43.1\% to 44.1\%.

Taken together, our analyses and results demonstrate the great potential of vocabulary projections as a framework for more principled research and development of dense retrieval models.

\section{Background}\label{sec:background}

In this work, we suggest a simple framework for interpreting dense retrieves, via projecting their representations to the vocabulary space. This is done using the (masked) language modeling head of their corresponding pretrained model. We begin by providing the relevant background.

\subsection{Masked Language Modeling}

Most language models based on encoder-only transformers \cite{vaswaniNIPS2017_7181} are pretrained using some variant of the masked language modeling (MLM) task \cite{devlin-etal-2019-bert,liu2019roberta,mpnet}, which involves masking some input tokens, and letting the model reconstruct them. 

Specifically, for an input sequence $x_1,...,x_n$, the transformer encoder is applied to output contextualized token representations $\bm{h}_1,...,\bm{h}_n\in\mathbb{R}^d$.
Then, to predict the missing tokens, an MLM head is applied to their contextualized representations. The MLM head is a function that takes a vector $\bm{h}\in\mathbb{R}^d$ as input and returns a distribution $P$ over the model's vocabulary $\mathcal{V}$, defined as follows:
\begin{equation}\label{eq:mlm}
\begin{split}
    \text{MLM-Head}(\bm{h})[i] = \frac{\exp(\bm{v}_i^\top g(\bm{h}))}{\sum_{j\in \mathcal{V}} \exp(\bm{v}_j^\top g(\bm{h}))} 
\end{split}
\end{equation}
$g:\mathbb{R}^d\rightarrow \mathbb{R}^d$ is a potentially non-linear function (\textit{e.g.}, a fully connected layer followed by a LayerNorm for BERT; \citealt{devlin-etal-2019-bert}), and $\bm{v}_i\in\mathbb{R}^d$ corresponds to the \textit{static} embedding of the $i$-th item in the vocabulary.

\subsection{Dense Retrieval}

In dense retrieval, we are given a corpus of passages $\mathcal{C}=\{p_1,...,p_m\}$ and a query $q$ (\textit{e.g}., a question or a fact to check), and we wish to compute query and passage representations ($\bm{e}_q$ and $\bm{e}_p$, respectively) such that similarity in this space implies high relevance of a passage to the query.
Formally, let $\text{Enc}_Q$ be a query encoder and  $\text{Enc}_P$ a passage encoder. These encoders are mappings from the input text to a vector in $\mathbb{R}^d$, and are obtained by fine-tuning a given LLM. Specifically, they return a pooled version of the LLM contextualized embeddings (\textit{e.g.}, the \texttt{[CLS]} embedding or mean pooling).
We denote the embedding of the query and passage vectors as follows:
\begin{equation}\label{eq:dense_retrieval}
    \begin{split}
        \bm{e}_q &= \text{Enc}_Q(q) \\
        \bm{e}_p &= \text{Enc}_P(p) \\
        %
    \end{split}
\end{equation}
To fine-tune retrievers, a similarity measure $s(q,p)$ is defined (\textit{e.g.}, the dot-product between $\bm{e}_q$ and $\bm{e}_q$ or their cosine similarity)
and the model is trained in a contrastive manner to maximize retriever accuracy \cite{lee-etal-2019-latent,karpukhin-etal-2020-dense}. Importantly, in this process, the MLM head function does not change at all.

\section{Vocabulary Projections}\label{sec:framework}


We now describe our framework for projecting query and passage representations of dense retrievers to the vocabulary space. 
Given a dense retrieval model, we utilize the MLM head of the model it was initialized from to map from encoder output representations to distributions over the vocabulary (Eq.~\ref{eq:mlm}). For example, for DPR  \cite{karpukhin-etal-2020-dense} we take BERT's MLM head, as DPR was initialized from BERT. 
Given a query $q$, we use the query encoder $\text{Enc}_Q$ to obtain its representation $\bm{e}_q$ as in Eq.~\ref{eq:dense_retrieval}. Similarly, for a passage $p$ we apply the passage encoder $\text{Enc}_P$ to get $\bm{e}_p$. We then apply the MLM head as in Eq.~\eqref{eq:mlm} to obtain the vocabulary projection:
\begin{equation}\label{eq:latent_dist}
    \begin{split}
        Q &= \text{MLM-Head}(\bm{e}_q) \\
        P &= \text{MLM-Head}(\bm{e}_p)
    \end{split}
\end{equation}
Note that it is not clear a-priori that $Q$ and $P$ will be meaningful in any way, as the encoder model has been changed since pretraining, while the MLM-head function remains fixed. Moreover, the MLM function has not been trained to decode ``pooled'' sequence-level representations (\textit{i.e.}, the results of CLS or mean pooling) during pretraining. Despite this intuition, in this work we argue that $P$ and $Q$ are actually highly intuitive and can facilitate a better understanding of dense retrievers.



\begin{table*}[t]
\centering
\small
\begin{tabular}{@{}p{0.11\textwidth}p{0.18\textwidth}p{0.45\textwidth}p{0.18\textwidth}@{}}
\toprule
\textbf{Question} & \textbf{top-20 in $Q$} & \textbf{Passage} & \textbf{top-20 in $P$} \\
\midrule
where do the great \textcolor{applegreen}{lakes} meet the \textcolor{applegreen}{\textcolor{applegreen}{ocean}} (A: the saint lawrence \textcolor{blue}{river}) & \textcolor{applegreen}{lakes} lake shore \textcolor{applegreen}{ocean} confluence \textcolor{blue}{river} water \textcolor{blue}{north} \textcolor{blue}{canada} meet \textcolor{blue}{east} land rivers canoe sea \textcolor{blue}{border} \textcolor{blue}{michigan} connecting both shores & the great \textcolor{applegreen}{lakes} , also called the laurent \#\#ian great \textcolor{applegreen}{lakes} and the great \textcolor{applegreen}{lakes} of \textcolor{blue}{north} america , are a series of inter \#\#connected freshwater \textcolor{applegreen}{lakes} located primarily in the upper mid - \textcolor{blue}{east} region of \textcolor{blue}{north} america , on the \textcolor{blue}{canada} – united states \textcolor{blue}{border} , which connect to the atlantic \textcolor{applegreen}{ocean} through the saint lawrence \textcolor{blue}{river} . they consist of \textcolor{applegreen}{lakes} superior , \textcolor{blue}{michigan} , huron ... & \textcolor{applegreen}{lakes} lake the \textcolor{blue}{canada} great freshwater water region ontario these central \textcolor{blue}{river} rivers large basin core area erie all four
 \\
 \midrule
 \textcolor{applegreen}{southern} \textcolor{applegreen}{soul} was considered the \textcolor{applegreen}{sound} of what independent record label (A: \textcolor{blue}{motown}) & \textcolor{applegreen}{southern} \textcolor{blue}{music} label \textcolor{applegreen}{soul} \textcolor{blue}{motown} \textcolor{blue}{blues} nashville vinyl \textcolor{applegreen}{sound} independent labels country records genre dixie record released \textcolor{blue}{gospel} jazz south & \textcolor{applegreen}{soul} \textcolor{blue}{music} . the key sub \#\#gen \#\#res of \textcolor{applegreen}{soul} include the detroit ( \textcolor{blue}{motown} ) style , a rhythmic \textcolor{blue}{music} influenced by \textcolor{blue}{gospel} ; " deep \textcolor{applegreen}{soul} " and " \textcolor{applegreen}{southern} \textcolor{applegreen}{soul} " , driving , energetic \textcolor{applegreen}{soul} styles combining r \& b with \textcolor{applegreen}{southern} \textcolor{blue}{gospel} \textcolor{blue}{music} \textcolor{applegreen}{sound} ; ... which came out of the rhythm and \textcolor{blue}{blues} style ... & \textcolor{applegreen}{soul} \textcolor{blue}{music} jazz funk \textcolor{blue}{blues} rock musical fusion genre black pure classical genres pop \textcolor{applegreen}{southern} melody art like rich urban \\
 \midrule
 who sings \textcolor{applegreen}{does} \textcolor{applegreen}{he} \textcolor{applegreen}{love} me with \textcolor{applegreen}{re} \textcolor{applegreen}{\#\#ba} (A: linda davis)  & \textcolor{blue}{duet} \textcolor{blue}{song} \textcolor{applegreen}{love} \textcolor{blue}{music} solo \textcolor{applegreen}{re} \textcolor{applegreen}{he} motown me his " pa \textcolor{blue}{album} \textcolor{blue}{songs} honey reprise bobby i peggy blues & " \textcolor{applegreen}{does he love} you " is a \textcolor{blue}{song} written by sandy knox and billy st \#\#rit \#\#ch , and recorded as a \textcolor{blue}{duet} by american country \textcolor{blue}{music} artists \textcolor{applegreen}{re} \textcolor{applegreen}{\#\#ba} mc \#\#ent \#\#ire and linda davis ... & \textcolor{applegreen}{he} you him i it she his john we \textcolor{applegreen}{love} paul who me \textcolor{applegreen}{does} did yes why they how this \\
\bottomrule
\end{tabular}
\caption{Examples of questions and gold passages from the development set of Natural Questions, along with their 20 top-scored tokens in projections of DPR representations. \textcolor{applegreen}{Green tokens} represent the lexical overlap signal (\textit{i.e.}, tokens that appear in both the question and the passage). \textcolor{blue}{Blue tokens} represent query expansion (\textit{i.e.}, tokens that do not appear in the question but do appear in the passage).}
\label{tab:examples}
\end{table*}
\section{Experiment Setup}

To evaluate our framework and method quantitatively, we consider several dense retrieval models and datasets. 

\subsection{Models}

We now list the retrievers used to demonstrate our framework and method. All dense models share the same architecture and size (\textit{i.e.}, that of BERT-base; 110M parameters), and all were trained in a contrastive fashion with in-batch negatives---the prominent paradigm for training dense models \cite{lee-etal-2019-latent,karpukhin-etal-2020-dense,chang2020pretraining,qu-etal-2021-rocketqa,ram-etal-2022-learning,izacard2022unsupervised,gtr,spar}. For the analysis, we use DPR \cite{karpukhin-etal-2020-dense} and BERT \cite{devlin-etal-2019-bert} as its pretrained baseline. For the results of our method, we also use S-MPNet \cite{reimers-gurevych-2019-sentence} and Spider \cite{ram-etal-2022-learning}. Our sparse retrieval model is BM25 \cite{bm25}. We refer the reader to App.~\ref{app:models} for more details.

\subsection{Datasets} 

We follow prior work \cite{karpukhin-etal-2020-dense,ram-etal-2022-learning} and consider six common open-domain question answering (QA) datasets for the evaluation of our framework: Natural Questions (NQ; \citealt{kwiatkowski-etal-2019-natural}), TriviaQA \cite{joshi-etal-2017-triviaqa}, WebQuestions (WQ; \citealt{berant-etal-2013-semantic}), CuratedTREC (TREC; \citealt{curated2015}), SQuAD \cite{rajpurkar-etal-2016-squad} and EntityQuestions (EntityQs; \citealt{sciavolino-etal-2021-simple}). We also consider the BEIR \cite{beir} and the MTEB \cite{mteb} benchmarks.

\subsection{Implementation Details}\label{sec:settings}

Our code is based on the official repository of DPR \cite{karpukhin-etal-2020-dense}, built on Hugging Face Transformers \cite{wolf-etal-2020-transformers}.


For the six QA datasets, we use the Wikipedia corpus standardized by \citet{karpukhin-etal-2020-dense}, which contains roughly 21 million passages of a hundred words each.
For dense retrieval over this corpus, we
apply exact search using FAISS \cite{faiss2021}. For sparse retrieval we use Pyserini \cite{pyserini}.
\section{Analyzing Dense Retrievers via Vocabulary Projections}\label{sec:analysis}

In Section~\ref{sec:framework}, we introduce a new framework for interpreting representations produced by dense retrievers.
Next, we describe empirical findings that shed new light on what is encoded in these representations. Via vocabulary projections, we draw connections between dense retrieval and well-known concepts from sparse retrieval like \textit{lexical overlap} (\S\ref{sec:lexical_overlap}), \textit{query prediction} (\S\ref{sec:query_prediction}) and \textit{query expansion} (\S\ref{sec:query_expansion}).

\subsection{The Dominance of Lexical Overlap}\label{sec:lexical_overlap}
\begin{figure}[t!]
\centering
\hspace*{-10pt}
\includegraphics[clip,width=\columnwidth]{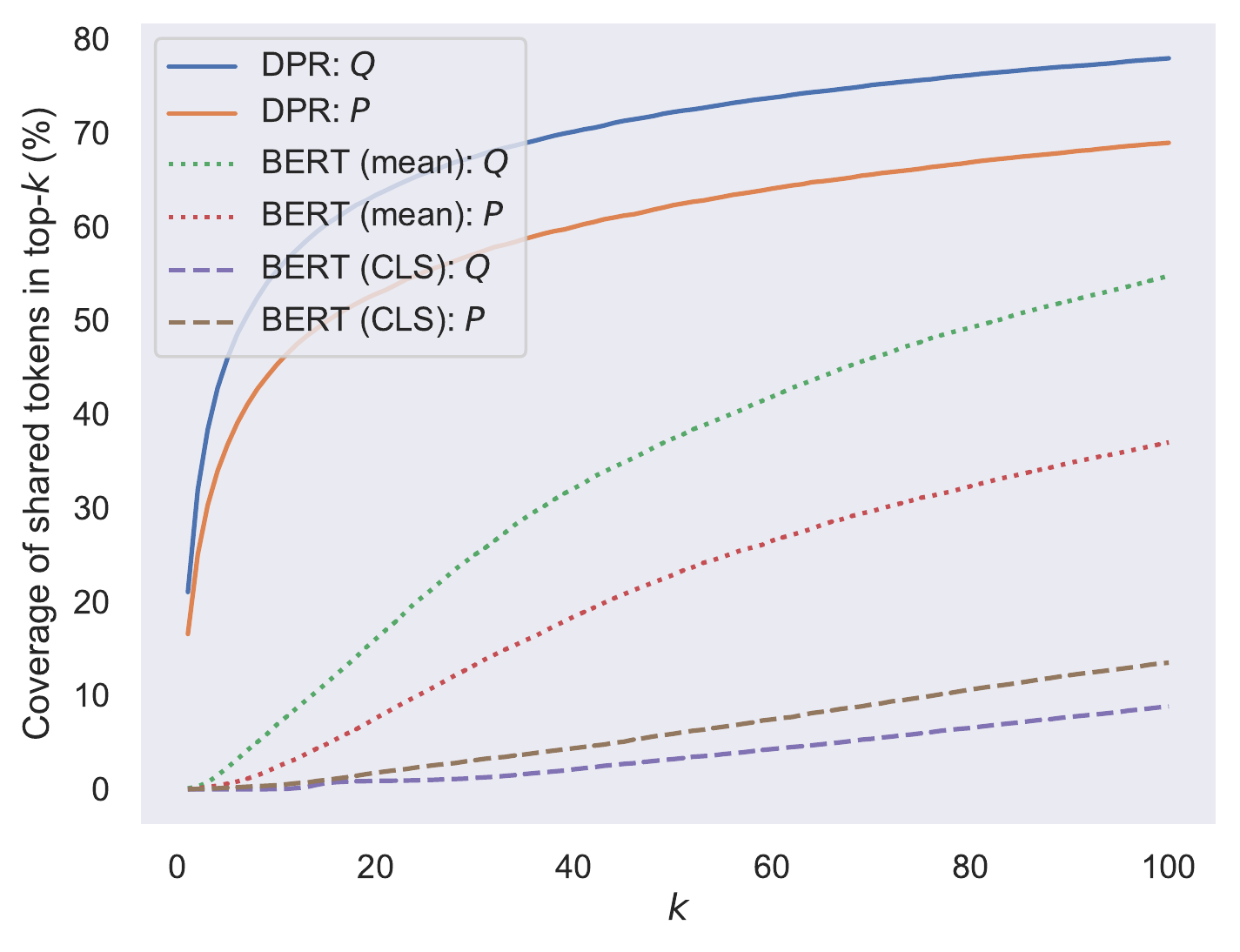}%
\caption{
The percentage of tokens shared by questions (from the development set of NQ) and their gold passages  (\textit{i.e.}, the lexical overlap signal) that are covered by the top-$k$ tokens of question vocabulary projection $Q$ and passage vocabulary projection $P$ as a function of $k$. Stop words and punctuation marks are excluded from this analysis.
}\label{fig:shared-tokens-coverage}
\end{figure}

Tokens shared by questions and their corresponding gold passages constitute the \textit{lexical overlap} signal in retrieval, used by sparse models like BM25. We start by asking: \textit{how prominent are they in vocabulary projections?} Figure~\ref{fig:shared-tokens-coverage}  illustrates the coverage of these tokens in $Q$ and $P$ for DPR after training, compared to its initialization before training (\textit{i.e.}, BERT with mean or CLS pooling). 
In other words, for each $k$ we check what is the percentage of shared tokens ranked in the top-$k$ of $Q$ and $P$.
Results suggest that after training, the model learns to rank shared tokens much higher than before. Concretely, 63\% and 53\% of the shared tokens appear in the top-20 tokens of $Q$ and $P$ respectively, compared to only 16\% and 8\% in BERT (\textit{i.e.}, before training). These numbers increase to 78\% and 69\% of the shared tokens that appear in the top-100 tokens of $Q$ and $P$. In addition, we observed that for 71\% of the questions, the top-scored token in $Q$ appears in both the question and the passage (App.~\ref{app:token_analysis}). 
These findings suggest that even for dense retrievers---which do not operate at the lexical level---lexical overlap remains a highly dominant signal.

\subsection{Passage Encoders as Query Prediction}\label{sec:query_prediction}

Our next analysis concerns the role of \textit{passage encoders}. 
In \S\ref{sec:lexical_overlap}, we show that tokens shared by the question and its gold passage are ranked high in both $Q$ and $P$. However, passages contain many tokens, and the shared tokens constitute only a small fraction of all tokens.
We hypothesize that out of passage tokens, \textit{those that are likely to appear in relevant questions receive higher scores in $P$ than others}. If this indeed the case, it implies that passage encoders implicitly learn to \textit{predict} which of the passage tokens will appear in relevant questions.
To test our hypothesis, we analyze the ranks of question and passage tokens in passage vocabulary projections, $P$. Formally, let $\mathcal{T}_q$ and $\mathcal{T}_p$ be the \textit{sets} of tokens in a question $q$ and its gold passage $p$, respectively.
Table~\ref{tab:token-mrr} shows the token-level mean reciprocal rank (MRR) of these sets in $P$. 
We observe that tokens shared by $q$ and $p$ (i.e., $\mathcal{T}_q\cap \mathcal{T}_p$) are ranked significantly higher than other passage tokens (\textit{i.e.}, $\mathcal{T}_p$). For example, in DPR the MRR of shared tokens is 26.1, while that of other passage tokens is only 3.0.
In addition, the MRR of shared tokens in BERT is only 1.4.
These findings support our claim that tokens that appear in relevant questions are ranked higher than others, and that this behavior is acquired during fine-tuning. 

\begin{table}[t!]
\footnotesize
\centering
\begin{tabular}{llcc}
\toprule
& & \multicolumn{2}{c}{\textbf{Token-Level MRR in $P$}} \\
\cmidrule(lr){3-4} & & ~~~DPR~~~ & BERT (mean) 
\\
\midrule
Passage tokens & $\mathcal{T}_p$ & ~~3.0 & ~~0.5 \\
Question tokens & $\mathcal{T}_q$ & 17.3 & ~~1.0 \\
Shared tokens & $\mathcal{T}_q\cap \mathcal{T}_p$ & 26.1 & ~~1.4 \\
\bottomrule
\end{tabular}
\caption{An analysis of token-level MRR (in \%) in \textbf{passage} vocabulary projections $P$ on the development set of NQ. For a question $q$ and its gold positive passage $p$, $\mathcal{T}_q$ and $\mathcal{T}_p$ are the corresponding sets of tokens, excluding stop words and punctuations. For a set $\mathcal{T}$, we report $\frac{1}{|\mathcal{T}|}\sum_{t\in\mathcal{T}}\frac{1}{\text{rank}_P(t)}$.
}
\label{tab:token-mrr}
\end{table}

\subsection{Query Encoders Implement Query Expansion}\label{sec:query_expansion}

\begin{figure}[t!]
\centering
\hspace*{-10pt}
\includegraphics[clip,width=\columnwidth]{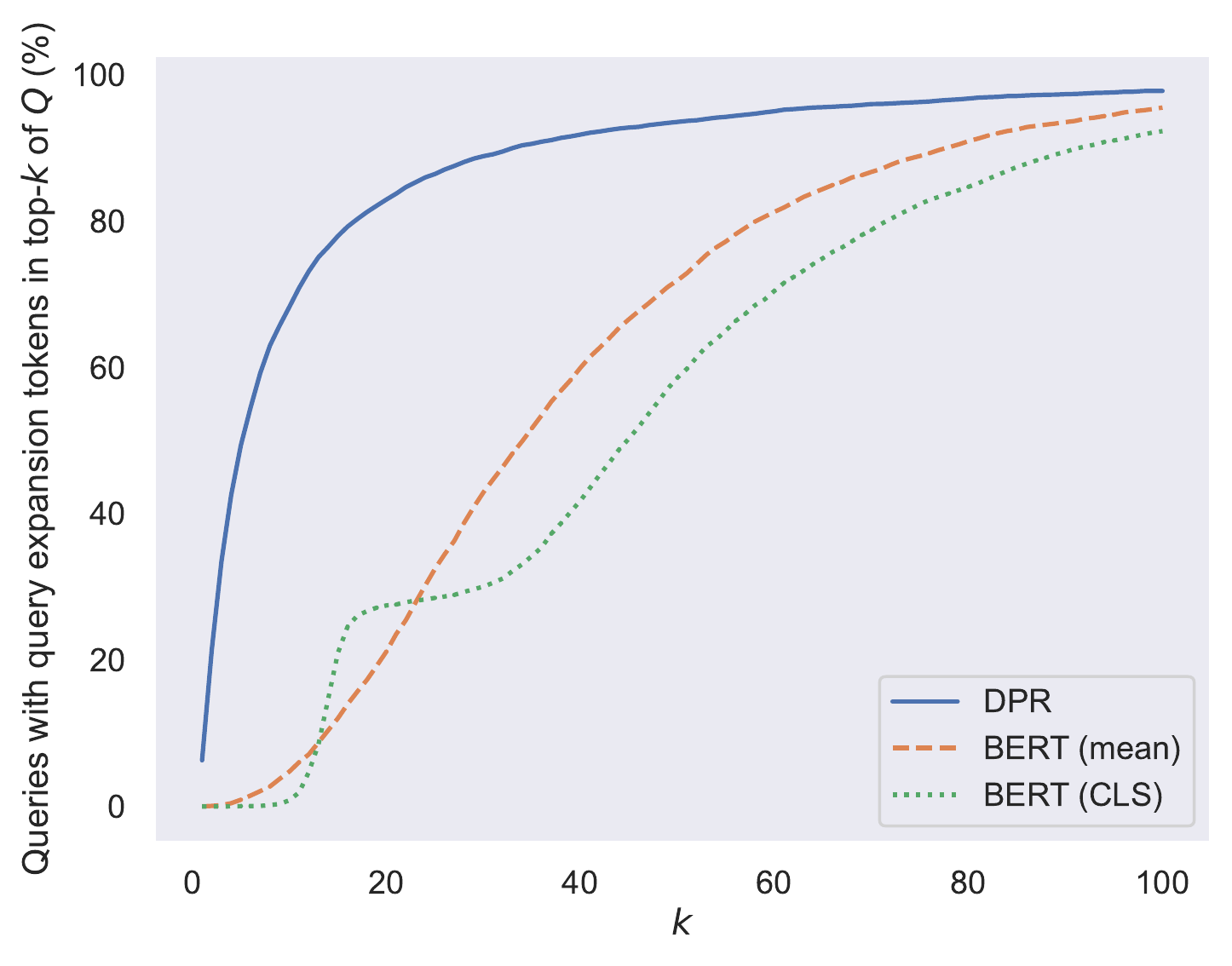}%
\caption{
The percentage of questions from the (entire) development set of NQ with at least one query expansion token (\textit{i.e.}, a token that appears in the question's gold passage but not in the question itself) in the top-$k$ of the question vocabulary projection $Q$, as a function of $k$. Stop words and punctuation marks do not count as query expansion tokens.
}\label{fig:query-expansion}
\end{figure}

To overcome the ``vocabulary mismatch'' problem (\textit{i.e.}, when question-document pairs are semantically relevant, but lack significant lexical overlap), \textit{query expansion} methods have been studied extensively \cite{relevance-feedback,qexpansion1,qexpansion2,mao-etal-2021-generation}. The main idea is to expand the query with additional terms that will better guide the retrieval process. 
We define a token as a query expansion if it does not appear in the query itself but does appear in the query projection $Q$, and also in the gold passage of that query $p$ (excluding stop words and punctuation marks). 
Figure~\ref{fig:query-expansion} shows the percentage of queries with at least one query expansion token in the top-$k$ as a function of $k$ for DPR and the BERT baseline (\textit{i.e.}, before DPR training). We observe that after training, the model promotes query expansion tokens to higher ranks than before. In addition, we found that almost 14\% of the tokens in the top-5 of $Q$ are query expansion tokens (\textit{cf}. App~\ref{app:token_analysis}).

We note that there are two interesting classes of query expansion tokens: (1) synonyms of question tokens, as well as tokens that share similar semantics with tokens in $q$ (\textit{e.g.}, ``\textcolor{blue}{michigan}'' in the first example of Table~\ref{tab:examples}). (2) ``answer tokens'' which contain the answer to the query (\textit{e.g.}, ``\textcolor{blue}{motown}'' in the second example of Table~\ref{tab:examples}). The presence  of such tokens may suggest the model already ``knows'' the answer to the given question, either from pretraining or from similar questions seen during training \cite{lewis-etal-2021-question}.

Given these findings, we conjecture that the model ``uses'' these query expansion tokens to introduce a semantic signal to the retrieval process.

\begin{figure*}[t!]
\centering
\vspace{-20pt}
\hspace*{-40pt}
\includegraphics[width=1.2\textwidth]{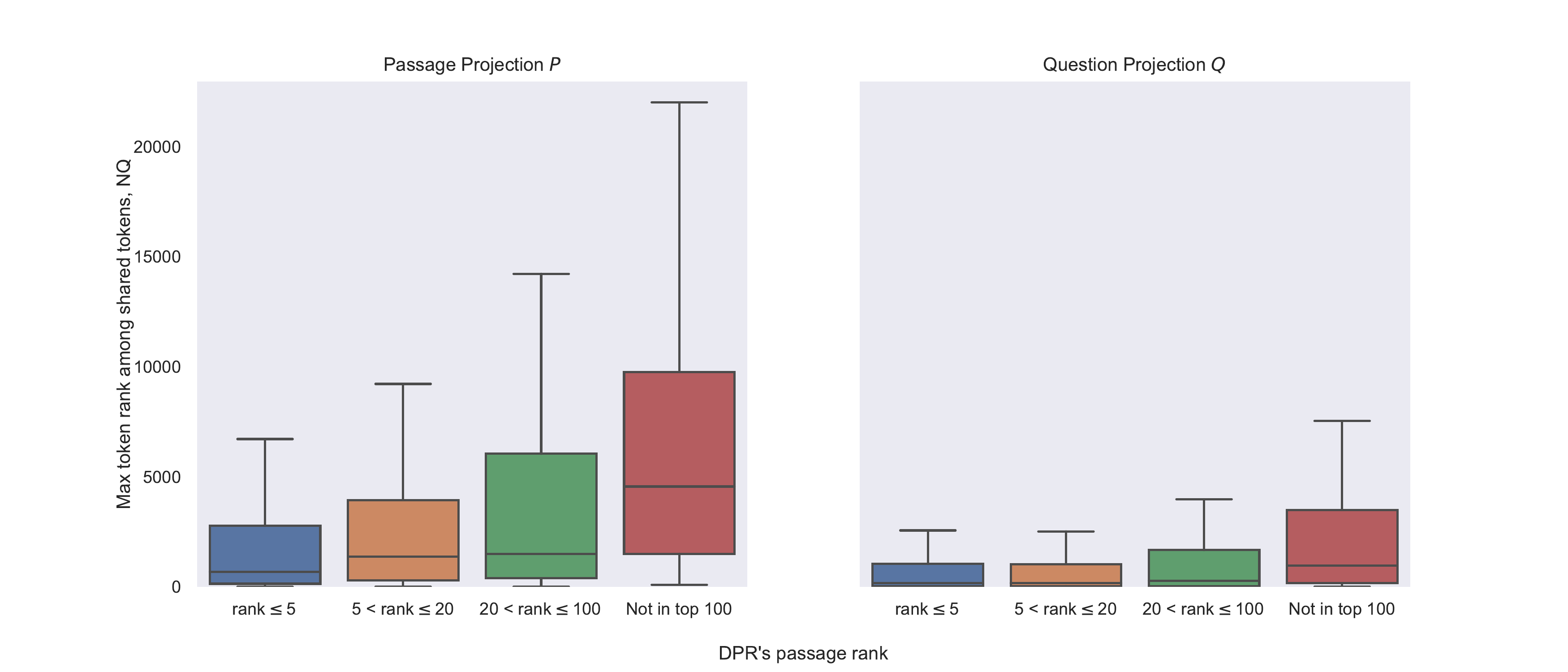}
\vspace{-20pt}
\caption{
An analysis of \textit{token amnesia}. 
We consider questions for which BM25 retrieves a correct passage (\textit{i.e.}, a passage that contains the answer) in its top-5, and analyze what ranks were assigned to tokens \textit{shared} by the question and the passage in the passage vocabulary projection $P$ (left) and question vocabulary projection $Q$ (right). We plot the maximal token rank as a function of the rank assigned to the correct passage by DPR.
}
\label{fig:failure}
\end{figure*}

\newcommand{\eqlex}{\bm{e}_q^\text{lex}}
\newcommand{\eplex}{\bm{e}_p^\text{lex}}
\newcommand{\exlex}{\bm{e}_x^\text{lex}}
\newcommand{\tokenrep}{\bm{e}_t}
\newcommand{\singleton}[1]{\bm{s}_{#1}}
\newcommand{\singletonh}[1]{\hat{\bm{s}}_{#1}}
\newcommand{\ignore}[1]{}

\section{Token Amnesia}\label{sec:token_forgetfulness}


The analysis in Section~\ref{sec:analysis} shows that vocabulary projections of passages (\textit{i.e.}, $P$) predict which of the input tokens are likely to appear in relevant questions. 
However, in some cases these predictions utterly fail.
For example, in Figure~\ref{fig:intro_two_examples} the token ``\textit{michael}'' is missing from the top-$k$ of the passage projection $P$.
We refer to such cases as \textit{token amnesia}.
Here we ask, \textit{do these failure in query prediction hurt retrieval?} 

Next, we demonstrate that token amnesia indeed correlates with well-known failures of dense retrievers (\S\ref{sec:token_forgetfulness_analysis}). To overcome this issue, we suggest a lexical enrichment procedure for dense representations (\S\ref{sec:token_forget_method}) and demonstrate its effectiveness on downstream retrieval performance (\S\ref{sec:token_forget_results}).

\subsection{Token Amnesia is Correlated with Retriever Failures}\label{sec:token_forgetfulness_analysis} 

Dense retrievers have shown difficulties in \textit{out-of-domain} settings \cite{sciavolino-etal-2021-simple,beir}, where even sparse models like BM25 significantly outperform them.
We now offer an intuitive explanation to these failures via token amnesia. We focus on setups where BM25 outperforms dense models and ask: \textit{why do dense retrievers fail to model lexical overlap signals?} To answer this question, we consider subsets of NQ and EntityQs where BM25 is able to retrieve the correct passage in its top-5 results. We focus on these subsets as they contain significant lexical overlap between questions and passages (by definition, as BM25 successfully retrieved the correct passage).
Let $q$ be a question and $p$ the passage retrieved by BM25 for $q$, and $Q$ and $P$ be their corresponding vocabulary projections for some dense retriever. 
Also, let $\mathcal{T}\subseteq\mathcal{V}$ be the set of tokens that appear in both $q$ and $p$ (excluding stop words).
Figure~\ref{fig:failure} shows the maximum (\textit{i.e.}, lowest) rank of tokens from $\mathcal{T}$ in the distributions $P$ (left) and $Q$ (right) as a function of whether DPR is able to retrieve this passage (\textit{i.e.}, the rank of $p$ in the retrieval results of DPR). Indeed, the median max-rank over questions for which DPR succeeds to fetch $p$ in its top-5 results (blue box) is much lower than that of questions for which DPR fails to retrieve the passage (red box).
As expected (due to the fact that questions contain less tokens than passages), the ranks of shared tokens in question projections $Q$ are much higher. However, the trend is present in $Q$ as well. Additional figures (for EntityQs; as well as median ranks instead of max ranks) are given in App.~\ref{app:token-amnesia}. 


Overall, these findings indicate a correlation between token amnesia and failures of DPR. 
Next, we introduce a method to address token amnesia in dense retrievers, via lexical enrichment of dense representations. 

\subsection{Method: Lexical Enrichment}\label{sec:token_forget_method}

As suggested by the analysis in \S\ref{sec:token_forgetfulness_analysis}, dense retrievers have the tendency to ignore some of their input tokens. We now leverage this insight to improve these models. We refer to our method as \textit{lexical enrichment} (LE) because it enriches text encodings with specific lexical items.

Intuitively, a natural remedy to the ``token amnesia'' problem is to change the retriever encoding such that {\em it does} include these tokens. For example, assume the query $q$ is ``\textit{Where was Michael Jack born?}'' and the corresponding passage $p$ contains the text ``\textit{Michael Jack was born in
Folkestone, England}''. According to Figure~\ref{fig:intro_two_examples}, the token ``\textit{michael}'' is ranked relatively low in $P$, and DPR fails to retrieve the correct passage $p$. We would like to modify the passage representation $\bm{e}_p$ and get an \textit{enriched} version $\bm{e}'_p$ that does have this token in its top-$k$ projected tokens, while keeping most of the other projected tokens intact. This is our goal in LE, and we next describe the approach. 
We focus on enrichment of passage representations, as query enrichment works similarly. We first explain how to enrich representations with a single token, and then extend the process to multiple tokens.

\paragraph{Single-Token Enrichment}
Assume we want to enrich a passage representation $\bm{e}_p$ with a token $t$ (\textit{e.g.}, $t=``$\textit{michael}'' in the above example). If there were no other words in the passage, we'd simply want to find an embedding such that feeding it into the MLM would produce $t$ as the top token.\footnote{Note that feeding the token input embedding $\bm{v}_t$ does not necessarily  produce $t$ as the top token, as the MLM head applies a non-linear function $g$ (Eq.~\ref{eq:mlm}).} We refer to this embedding as the \textit{single-token enrichment} of $t$, denote it by $\singleton{t}$ and define it as:\footnote{This is equivalent to the cross-entropy loss between a one-hot vector on $t$ and the output distribution $MLM(\singletonh{})$.}
\begin{equation}\label{eq:token-embeddings}
\singleton{t} = \argmax_{\singletonh{}}\ \  \log \text{MLM-Head}(\singletonh{})[t]
\end{equation}
In order to approximately solve the optimization problem in Eq.~\ref{eq:token-embeddings} for each $t$ in the vocabulary, we use Adam with a learning rate of 0.01.\footnote{For S-MPNet, we used a learning rate of $10^{-3}$.}
We stop when a (cross-entropy) loss threshold of 0.1 is reached for all tokens. We then apply whitening \cite{whitening}, which was proven effective for dense retrieval.

\begin{table*}[t]
\centering
\small
\begin{tabular}{lcccccccc}
\toprule
\multirow{2.6}{50pt}{\textbf{Model}} & \multirow{2.6}{5pt}{$\lambda$} & \textbf{~BEIR~} & \textbf{~MTEB~} &
\textbf{EntityQs} & \textbf{TriviaQA} & \textbf{~~WQ~~} & \textbf{~~TREC~~} & \textbf{SQuAD}  \\
\cmidrule(lr){3-4} \cmidrule(lr){5-9}
&& \multicolumn{2}{c}{\textbf{nDCG@10}} & \multicolumn{5}{c}{\textbf{Top-20 retrieval accuracy}} \\
\midrule
BM25 & - & 42.9 & 42.3 & \underline{71.4} & 76.4 & 62.4 & 81.1 & \underline{71.2} \\
BM25 (BERT/MPNet Tokens) & - & 41.6 & 41.7 & 66.2 & 75.8 & 62.1 & 79.3 & 70.0 \\
\midrule
DPR  & - & 21.4 & 22.4 & 49.7 & 69.0 & 68.8 & 85.9 & 48.9 \\
DPR + LE & 5.0 & \textbf{26.4} & \textbf{27.6} & \textbf{65.4} & \textbf{75.3} & \textbf{73.2} & \textbf{87.9} & \textbf{59.7} \\
\midrule
S-MPNet & - & 43.1 & 44.6 & 57.6 & 77.6 & 73.9 & 90.2 & 65.5 \\
S-MPNet + LE & 0.5 & \textbf{\underline{44.1}} & \textbf{\underline{45.7}} & \textbf{68.5} & \textbf{\underline{78.9}} & \textbf{\underline{74.5}} & \textbf{\underline{90.4}} & \textbf{69.0} \\
\midrule
Spider & - & 27.4 & 26.4 & 66.3 & 75.8 & 65.9 & 82.6 & 61.0  \\
Spider + LE & 3.0 & \textbf{29.5} & \textbf{28.8} & \textbf{68.9} & \textbf{76.3} & \textbf{70.2} & \textbf{83.4} & 
\textbf{62.8} \\
\bottomrule
\end{tabular}
\caption{Retrieval results on BEIR, the retrieval cluster of MTEB and five open-domain QA datasets.
LE stands for \textit{lexical enrichment} (our method; \S\ref{sec:token_forget_method}), that enriches query and passage representation with lexical information.
$\lambda$ is defined in Eq.~\ref{eq:exlex}.
BM25 (BERT Vocabulary) refers to a model that operates over tokens from BERT's vocabulary, rather than words. For each model and dataset, we compare the enriched (LE) model with the original, and mark in bold the better one from the two. We underline the best overall model for each dataset. 
Results for each of the BEIR datasets are given in Table~\ref{tab:beir}.
Top-$\{1,5,100\}$ accuracy results are given in Tables~\ref{tab:zero-shot-k1},~\ref{tab:zero-shot-k5} \&~\ref{tab:zero-shot-k100}.
}
\label{tab:zero-shot}
\end{table*}

\paragraph{Multi-Token Enrichment}
Now suppose we have an input $x$ (either a question or a passage) and we'd like to enrich its representation with its tokens $x=[x_1,..,x_n]$, such that rare tokens are given higher weights than frequent ones (as in BM25). 
Then, we simply take its original representation $\bm{e}_x$ and add to it a weighted sum of the single-token enrichments (Eq.~\ref{eq:token-embeddings}). 
Namely, we define:
\begin{equation}\label{eq:exlex}
\begin{split}
    \exlex &= \frac{1}{n}\sum_{i=1}^n w_{x_i}\singleton{x_i} \\
    \bm{e}'_x  &= \bm{e}_x  +  \lambda\cdot\frac{\exlex}{||\exlex||}
\end{split}
\end{equation}
Here $\lambda$ is a hyper-parameter chosen via cross validation.
We use the inverse document frequency \cite{idf} of tokens as their weights:  $w_{x_i}=\text{IDF}(x_i)$.
 The relevance score is then defined on the enriched representations.

\ignore{
, we want to add to enrich
We consider an enrich
Enriched representations are defined by: 
\begin{equation}
\begin{split}
\bm{e}'_q &= \bm{e}_q + \lambda\cdot \eqlex \\
\bm{e}'_p &= \bm{e}_p + \lambda\cdot \eplex 
\end{split}
\end{equation}
where $\eqlex$ is a lexical representation of the question $q$ induced by the model, and similarly $\eplex$ for $p$. $\lambda$ is a hyper-parameter validated on the development set of NQ and EntityQs. The similarity score $s(q,p)$ is defined over the new representations $\bm{e}'_q$ and $\bm{e}'_p$.\footnote{The score is either dot product or cosine similarity, according to the original model.}

\paragraph{Projection-Aware Lexical Representations} We now turn to describe our implementation of the lexical representations, $\eqlex$ and $\eplex$. First, we build a static embedding $\tokenrep$ for each item $t$ in the model vocabulary $\mathcal{V}$. Ideally, we want $\tokenrep$ to be aligned with the model embedding space, so we take
\begin{equation}\label{eq:token-embeddings}
\begin{split}
\tokenrep &= \argmax_{\bm{e}}\ \  \log p_t^\bm{e} \\
p_t^\bm{e} &= \text{MLM}(\bm{e})_t
\end{split}
\end{equation}
Put differently, for each vocabulary item $t$ we take $\tokenrep$ to be a vector for which the output of the MLM head resembles a one-hot vector, peaked at $t$.

For an input $\bm{x}=[x_1,..,x_n]$ (either a question or a passage), we then define
\begin{equation}\label{eq:exlex}
\begin{split}
\exlex &= \frac{1}{n}\sum_{i=1}^n w_{x_i}\bm{e}_{x_i} \\
w_{x_i} &= \text{IDF}(x_i)
\end{split}
\end{equation}
Our use of inverse document frequency (IDF; \citealp{idf}) is inspired by BM25, which is known for its robustness across domains. To get our final lexical representation, we apply $\ell_2$-normalization on $\exlex$.
}

\subsection{Results}\label{sec:token_forget_results}
Our experiments demonstrate the effectiveness of our method for multiple models, especially in zero-shot settings. Table~\ref{tab:zero-shot} shows the results of several models with and without our enrichment method, LE. Additional results are given in App.~\ref{app:method}.  The results demonstrate the effectiveness of LE when added to all baseline models. 
Importantly, our method improves the performance of S-MPNet---the best base-sized model on the MTEB benchmark to date \cite{mteb}---on MTEB and BEIR by 1.1\% and 1.0\%, respectively.
When considering EntityQs (on which dense retrievers are known to struggle), we observe significant gains across all models,
and S-MPNet and Spider obtain higher accuracy than BM25 that operates on the same textual units (\textit{i.e.}, BM25 with BERT vocabulary). This finding indicates that they are able to integrate semantic information (from the original representation) with lexical signals.
Yet, vanilla BM25 is still better than LE models on EntityQs and SQuAD, which prompts further work on how to incorporate lexical signals in dense retrieval.
Overall, it is evident that LE improves retrieval accuracy compared to baseline models for all models and datasets (\textit{i.e.}, zero-shot setting). 


\begin{table*}[]
\small
\centering
\begin{tabular}{lllcccccccc}
\toprule
\multicolumn{1}{l}{\multirow{2.6}{*}{\textbf{Method}}} &
   &
  \multicolumn{4}{c}{\textbf{NQ (Dev Set)}} &
  \multicolumn{1}{c}{} &
  \multicolumn{4}{c}{\textbf{EntityQs (Dev Set)}} \\ \cmidrule(lr){3-6} \cmidrule(l){8-11} 
\multicolumn{1}{c}{} &
   &
  \textbf{Top-1} &
  \textbf{Top-5} &
  \textbf{Top-20} &
  \textbf{Top-100} &
   &
  \textbf{Top-1} &
  \textbf{Top-5} &
  \textbf{Top-20} &
  \textbf{Top-100} \\ \midrule
DPR &  & 44.9 & 66.8          & 78.1 & 85.0          &  & 24.0   & 38.4 & 50.4 & 63.5 \\
\midrule
DPR + LE &
   &
  44.4 &
  67.5 &
  \textbf{79.4} &
  \textbf{86.0} &
   &
  \textbf{38.3} &
  \textbf{54.0} &
  \textbf{65.2} &
  \textbf{76.1} \\
\quad \textit{No IDF}    &  & \textbf{45.1} & 67.3          & 78.5 & 85.4        &  & 32.0   & 46.4 & 57.7 & 69.6 \\
\quad \textit{BERT embedding matrix} &  & 44.8          & \textbf{67.6} & 79.1 & 85.6        &  & 34.6 & 50.3 & 61.8 & 72.8 \\
\quad \textit{No whitening}      &  & 44.1          & 66.3          & 78.7 & 85.2        &  & 34.6 & 49.7 & 61.4 & 72.9 \\
\quad \textit{No $\ell_2$ normalization}  &  & 43.9          & 66.8          & 79.2 & \textbf{86.0} &  & 35.5 & 51.3 & 63.0   & 74.6 \\ \bottomrule
\end{tabular}
\caption{Ablation study on the development set of Natrual Questions and Entity Questions. DPR + LE is our lexical enrichment method applied on DPR. \textit{No IDF} removes the IDF weights in Eq.~\ref{eq:exlex} (i.e., mean pooling). \textit{BERT embedding matrix} replaces single-token enrichment $\bm{s}_t$ as defined in Eq.~\ref{eq:token-embeddings} with the static token embeddings of BERT, $\bm{v}_t$ (Eq.~\ref{eq:mlm}). \textit{No whitening} removes whitening transformation. \textit{No $\ell_2$ normalization} removes the normalization of $\exlex$.
}\label{tab:ablation}
\end{table*}
\subsection{Ablation Study}\label{sec:ablations}

We carry an ablation study to test our design choices from \S\ref{sec:token_forget_method}. We evaluate four elements of our method: (1) The use of IDF to highlight rare tokens, (2) Our approach for deriving single-token representations, (3) The use of whitening, and (4) The use of unit normalization.

\paragraph{IDF} In our method, we create lexical representations of questions and passages, $\exlex$.
These lexical representations are the average of token embeddings, each multiplied by its token's IDF. We validate that IDF is indeed necessary -- Table~\ref{tab:ablation} demonstrates that setting $w_{x_i}=1$ in Eq.~\ref{eq:exlex} leads to a significant degradation in performance on EntityQs. For example, top-20 retrieval accuracy drops from 65.2\% to 57.7\%.

\paragraph{Single-Token Enrichment} Eq.~\ref{eq:token-embeddings} defines our single-token enrichment: for each item in the vocabulary $v\in\mathcal{V}$, we find an embedding which gives a one-hot vector peaked at $v$ when fed to the MLM head. 
We confirm that this is necessary by replacing Eq.~\ref{eq:token-embeddings} with the static embeddings of the pretrained model (e.g., BERT in the case of DPR). We find that our approach significantly improves over BERT's embeddings on EntityQs (\textit{e.g.}, the margin in top-20 accuracy is 3.4\%).

\paragraph{Whitening \& Normalization} Last, we experiment with removing the whitening and $\ell_2$ normalization. It is evident that they are both necessary, as removing either of them causes a dramatic drop in performance (3.8\% and 2.2\% in top-20 accuracy on EntityQs, respectively).

\section{Related Work}

Projecting representations and model parameters to the vocabulary space has been studied previously mainly in the context of language models. The approach was initially explored by \citet{Nostalgebraist2020}. 
\citet{geva-etal-2021-transformer} showed that feed-forward layers in transformers can be regarded as key-value memories, where the value vectors induce distributions over the vocabulary. \citet{geva-etal-2022-transformer} view the token representations themselves as inducing such distributions, with feed-forward layers ``updating'' them. \citet{dar2022analyzing} suggest to project all transformer parameters to the vocabulary space. Dense retrieval models, however, do not have any language modeling objective during fine-tuning, yet we show that their representations can still be projected to the vocabulary.

Despite the wide success of dense retrievers recently, interpreting their representations remains under-explored. 
\citet{macavaney-etal-2022-abnirml} analyze neural retrieval models (not only dense retrievers) via diagnostic probes, testing characteristics like sensitivity to paraphrases, styles and factuality.
\citet{adolphs-etal-2022-decoding} decode the query representations of neural retrievers using a T5 decoder, and show how to ``move'' in representation space to decode better queries for retrieval.

Language models (and specifically MLMs) have been used for \textit{sparse retrieval} in the context of term-weighting and lexical expansion. For example, \citet{sparterm} and \citet{splade} learn such functions over BERT's vocabulary space.
We differ by showing that \textit{dense retrievers} implicitly operate in that space as well. Thus, these approaches may prove effective for dense models as well. While we focus in this work  on dense retrievers based on encoder-only models, our framework is easily extendable for retrievers based on autoregressive decoder-only (\textit{i.e.}, left-to-right) models like GPT \cite{gpt2,gpt3}, \textit{e.g.}, \citet{openai-embeddings} and \citet{sgpt}.

\section{Conclusion}

In this work, we explore projecting query and passage representations obtained by dense retrieval to the vocabulary space. 
We show that these projections facilitate a better understanding of the mechanisms underlying dense retrieval, as well as their failures. We also demonstrate how projections can help improve these models. This understanding is likely to help in improving retrievers, as our lexical enrichment approach demonstrates.

\section*{Limitations}

We point to several limitations of our work. First, our work considers a popular family of models referred to as ``dense retrievers'', but other approaches for retrieval include sparse retrievers \cite{bm25,sparterm,splade}, generative retrievers \cite{tay2022transformer,seal}, late-interaction models \cite{colbert}, \textit{inter alia}.
While our work draws interesting connections between dense and sparse retrieval, our main focus is on understanding and improving dense models.
Second, all three dense models we analyze are bidirectional and were trained in a contrastive fashion. While most dense retrievers indeed satisfy these properties, there are works that suggested other approaches, both in terms of other architectures  \cite{sgpt,openai-embeddings,gtr} and other training frameworks \cite{rag,izacard2022atlas}. 
Last, while our work introduces new ways to interpret and analyze dense retrieval models, we believe our work is the tip of the iceberg, and there is still much work to be done in order to gain a full understanding of these models.

\section*{Ethics Statement}
Retrieval systems have the potential to mitigate serious problems caused by language models, like factual inaccuracies. However, retrieval failures may lead to undesirable behavior of downstream models, like wrong answers in QA or incorrect generations for other tasks. Also, since retrieval models are based on pretrained language models, they may suffer from similar biases.

\section*{Acknowledgements}
We thank Ori Yoran, Yoav Levine, Yuval Kirstain, Mor Geva and the anonymous reviewers for their valuable feedback. This project was funded by the European Research Council (ERC) under the European Unions Horizon 2020 research and innovation programme (grant ERC HOLI 819080), the Blavatnik Fund, the Alon Scholarship, the Yandex Initiative for Machine Learning, Intel Corporation, ISRAEL SCIENCE FOUNDATION (grant No.\ 448/20), Open Philanthropy, and an Azrieli Foundation Early Career Faculty Fellowship.

\bibliography{anthology,custom}
\bibliographystyle{acl_natbib}

\appendix

\section{Models: Further Details}\label{app:models}

\paragraph{DPR} \cite{karpukhin-etal-2020-dense} is a  dense retriever that was trained on Natural Questions \cite{kwiatkowski-etal-2019-natural}. It was initialized from BERT-base \cite{devlin-etal-2019-bert}. Thus, we use the public pretrained MLM head of BERT-base to project DPR representations.

\paragraph{BERT} \cite{devlin-etal-2019-bert} We use BERT for dense retrieval, mainly as a baseline for DPR, as DPR was initialized from BERT. This allows us to track where behaviors we observe stem from: pretraining or retrieval fine-tuning.
We use both CLS and mean pooling for BERT. 

\paragraph{S-MPNet} is a supervised model trained for Sentence Transformers \cite{reimers-gurevych-2019-sentence} using many available datasets for retrieval, sentence similarity, \textit{inter alia}. It uses cosine similarity, rather than dot product, for relevance scores. It was initialized from MPNet-base \cite{mpnet}, and thus we use this model's MLM head.

\paragraph{Spider} \cite{ram-etal-2022-learning} is an unsupervised dense retriever trained using the \textit{recurring span retrieval} pretraining task. It was also initialized from BERT-base, and we therefore use the same MLM head for projection as the one used for DPR.

\paragraph{BM25} \cite{bm25} is a lexical model based on tf-idf. We use two variants of BM25: (1) vanilla BM25, and (2) BM25 over BERT/MPNet tokens (e.g., ``\textit{Reba}'' $\rightarrow$ ``\textit{re \#\#ba}'').\footnote{BERT and MPNet use essentially the same vocabulary, up to special tokens.} We consider this option to understand whether the advantages of BM25 stem from its use of different word units from the transformer models.

\section{Analysis: Further Results}\label{app:token_analysis}
\begin{figure}[t!]
\centering
\subfloat[Question projection $Q$]{%
  \includegraphics[clip,width=\columnwidth]{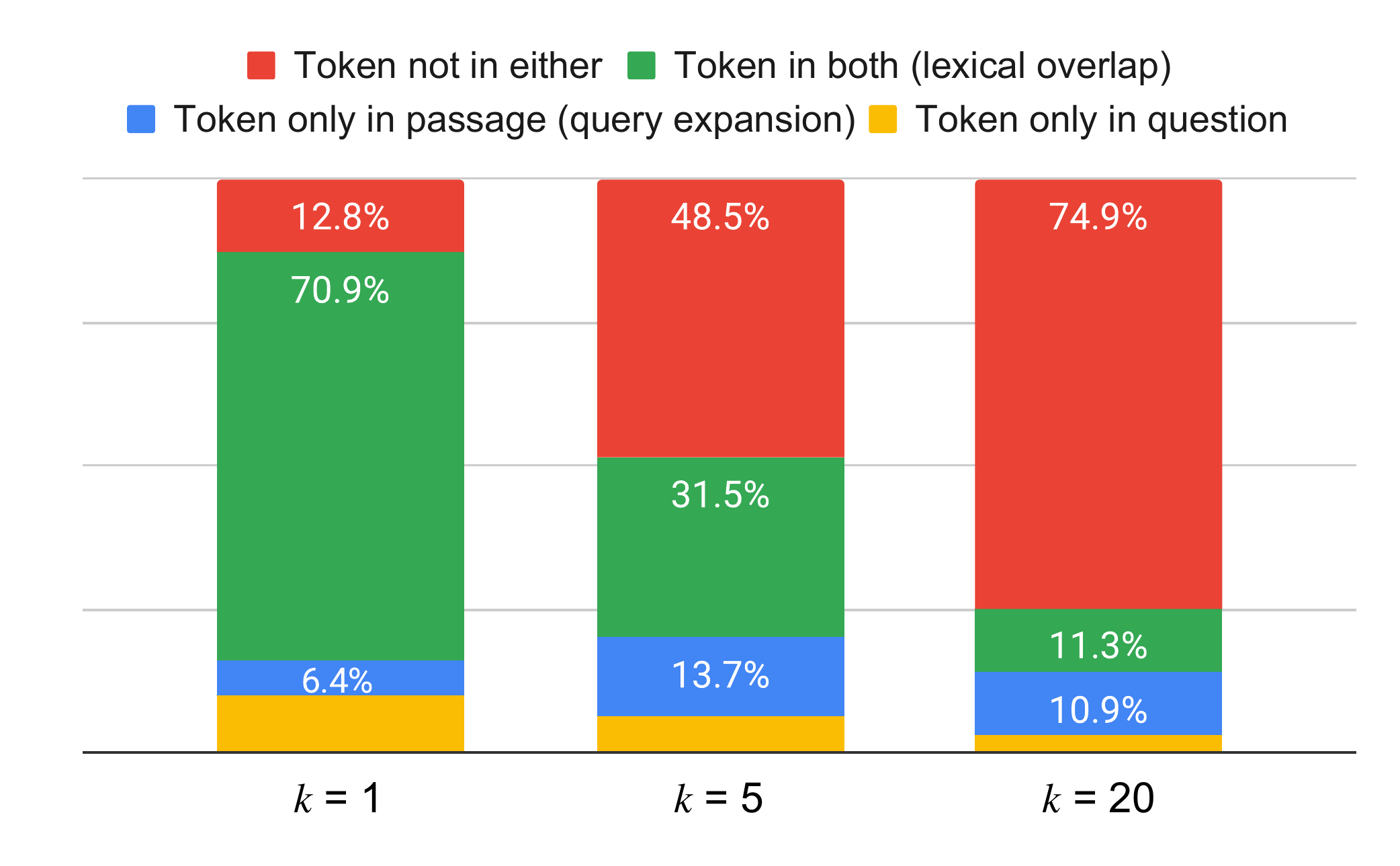}%
}

\subfloat[Passage projection $P$]{%
\includegraphics[clip,width=\columnwidth]{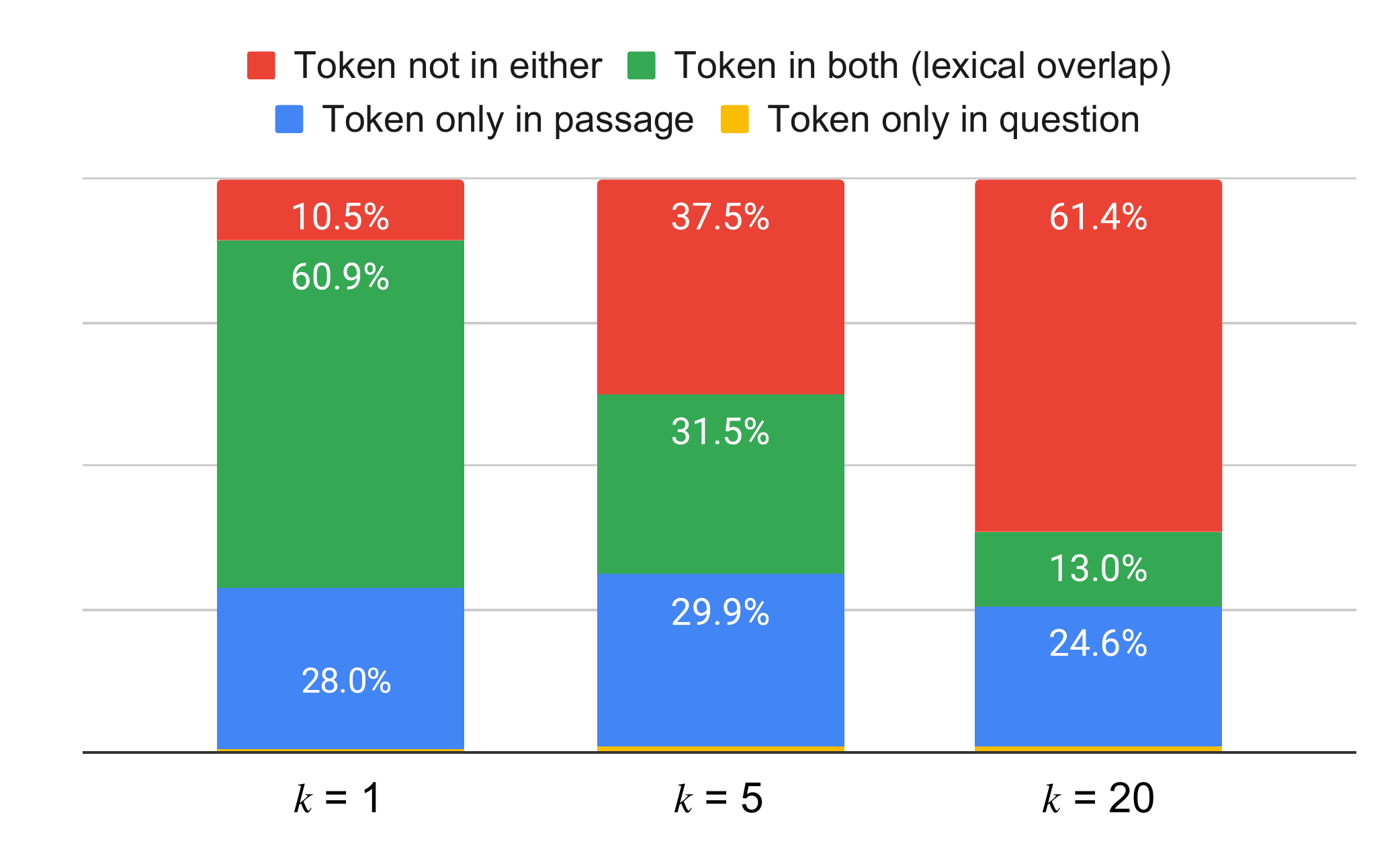}%
}
\caption{An analysis of the top-$k$ tokens in the vocabulary projection $Q$ (a) for questions from the development set of NQ and $P$ (b) for their corresponding gold passage of DPR. Specifically, we analyze what percentage of these top-$k$ tokens are present in the question and/or the passage for $k\in\{1,5,20\}$.
}\label{fig:stacked_Q_P}
\end{figure}

Figure~\ref{fig:stacked_Q_P} gives an analysis of the top-$k$ tokens in the question projection $Q$ and passage projection $P$.


\section{Token Amnesia: Further results}\label{app:token-amnesia}

\begin{figure*}[t!]
\centering
\vspace{-30pt}
\subfloat[Max rank among shared tokens, EntityQuestions]{%
  \includegraphics[clip,width=0.8\textwidth]{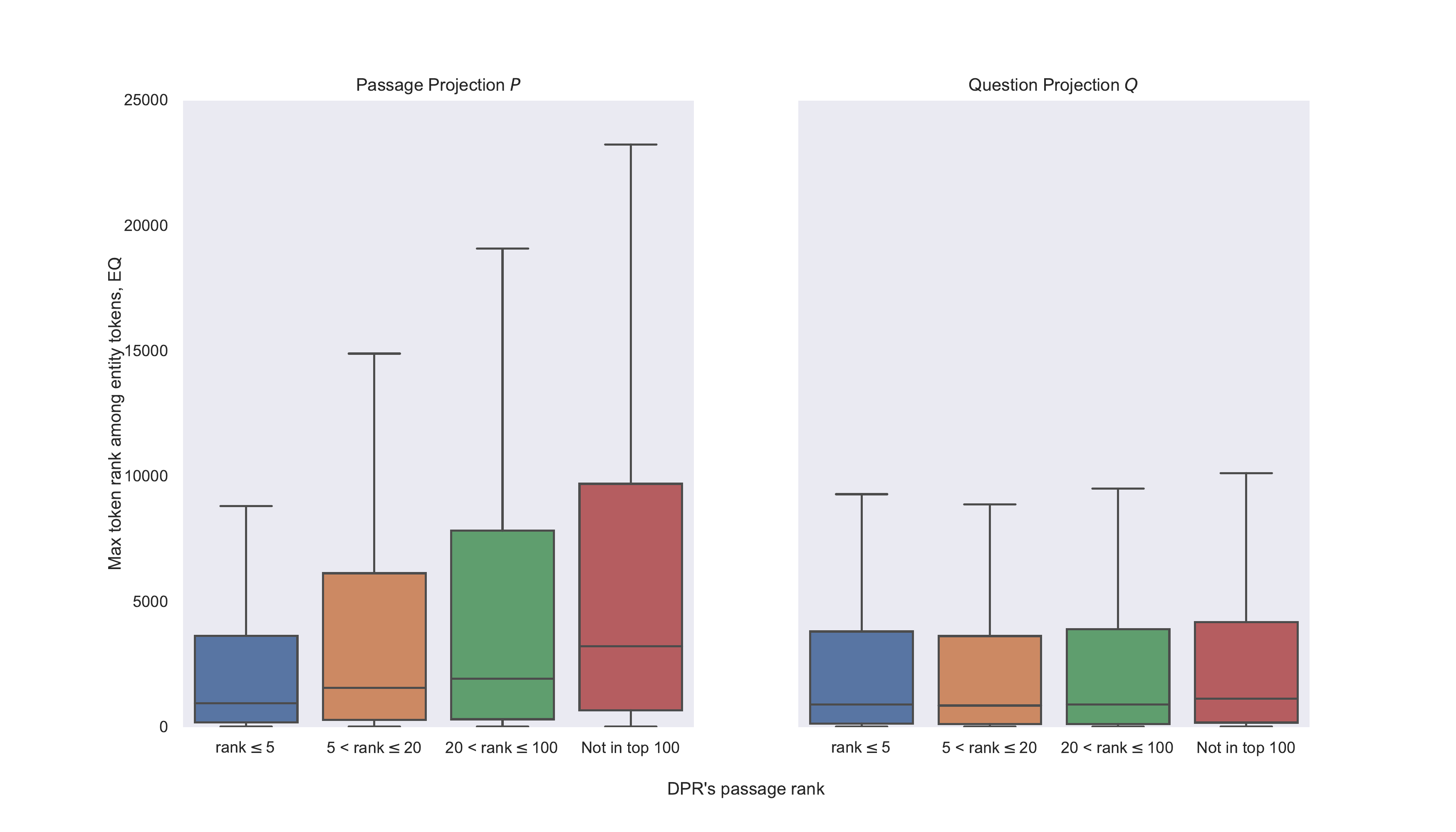}%
}

\vspace{-10pt}
\subfloat[Median rank among shared tokens, Natural Questions]{%
  \includegraphics[clip,width=0.8\textwidth]{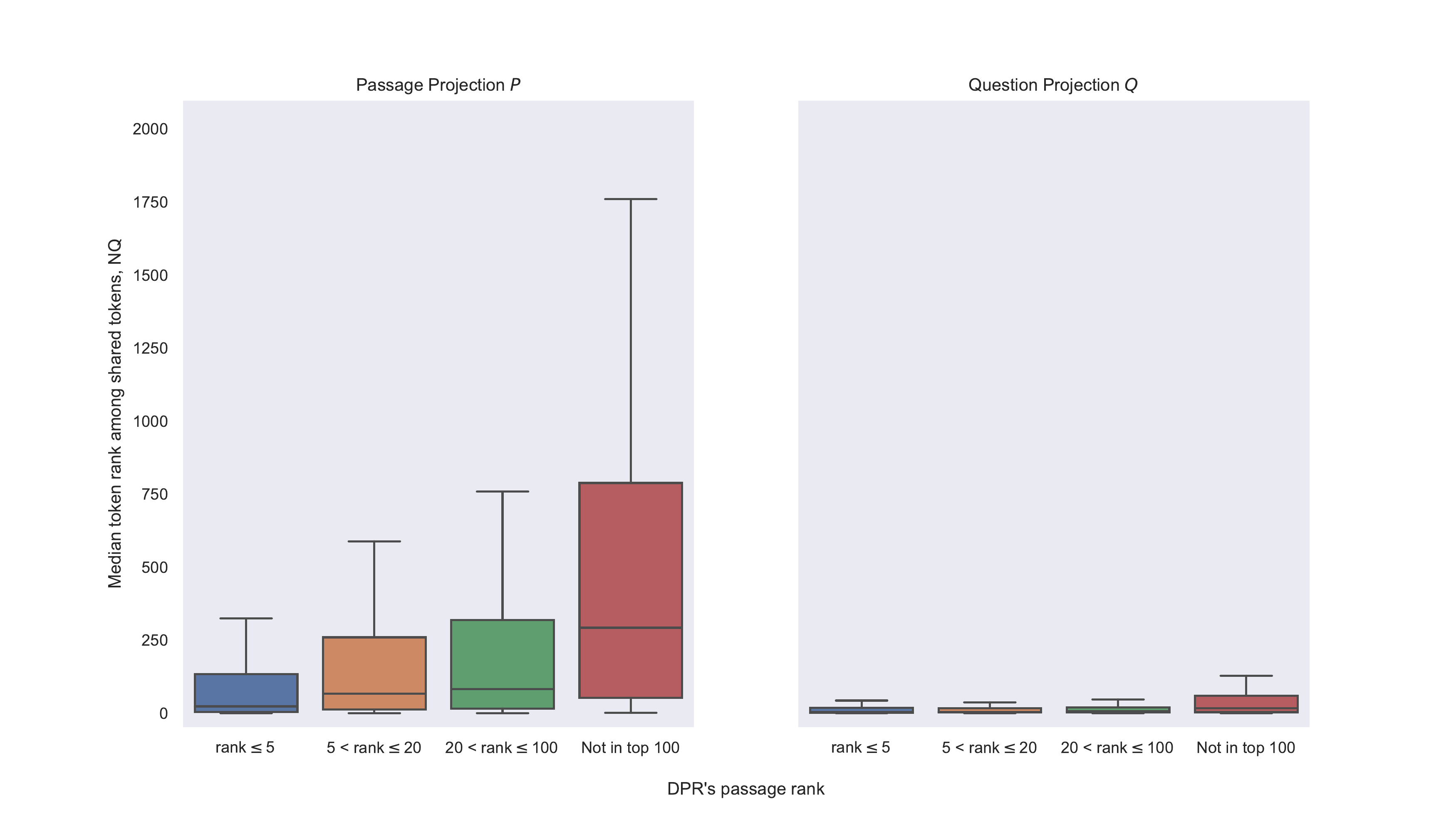}%
}

\vspace{-10pt}
\subfloat[Median rank among shared tokens, EntityQuestions]{%
  \includegraphics[clip,width=0.8\textwidth]{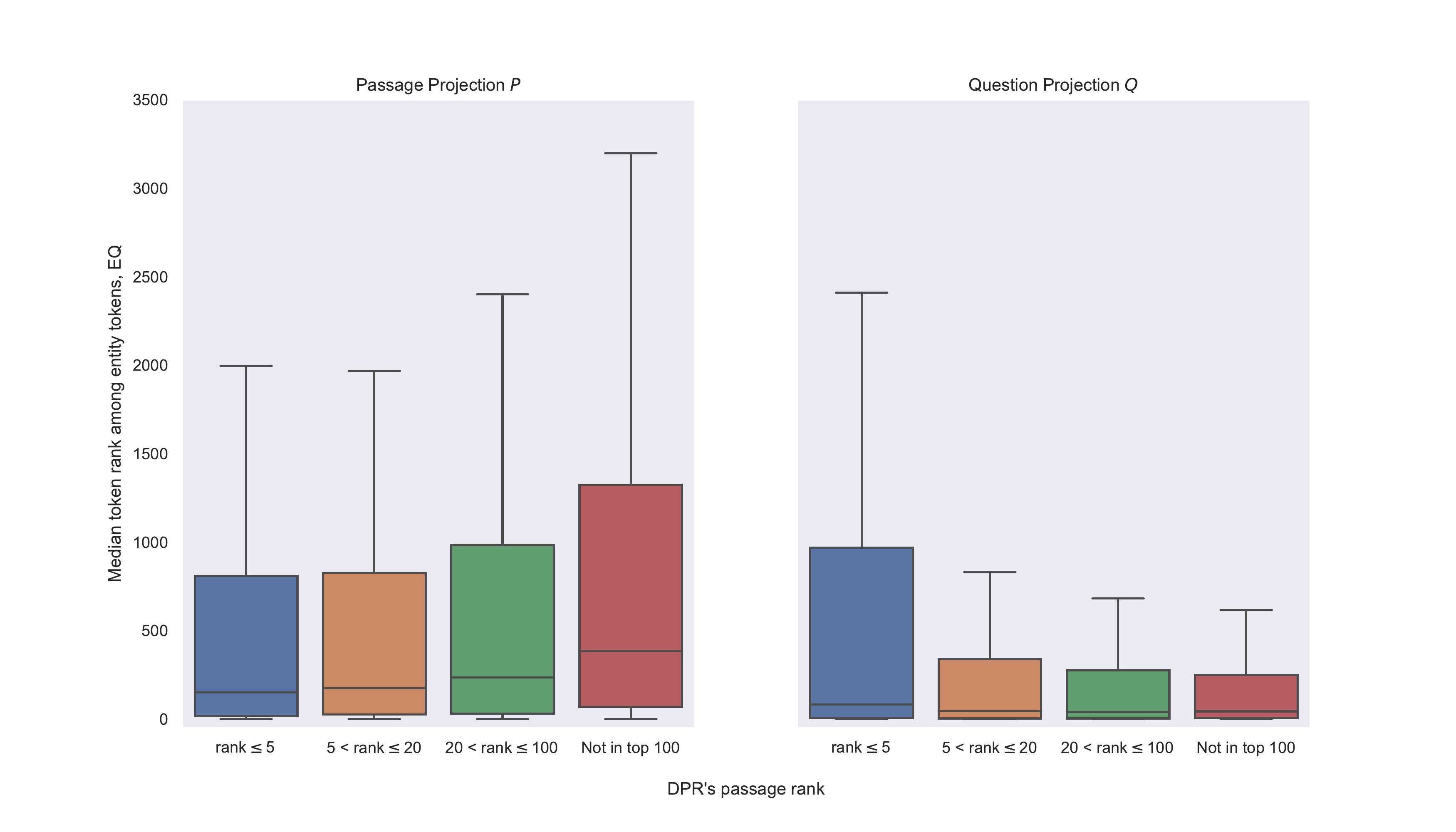}%
}

\caption{
Further analysis of \textit{token amnesia} (complementary to Figure~\ref{fig:failure}). 
We consider questions for which BM25 retrieves a correct passage (i.e., a passage that contains the answer) in its top-5, and analyze what ranks were assigned to tokens \textit{shared} by the question and the passage in the passage vocabulary projection $P$ (left) and question vocabulary projection $Q$ (right). We plot the max and median token rank as a function of the rank assigned to the correct passage by DPR, for Natural Questions (NQ) and EntityQuestions (EQ).
}
\label{fig:failure-app}
\end{figure*}

Figure~\ref{fig:failure-app} gives further analyses of token amnesia: It contains the results for EntityQuestions, as well as analysis of median ranks in addition to max ranks (complements Figure~\ref{fig:failure}).

\section{Lexical Enrichment: Further Results}\label{app:method}

\begin{table}[t!]
\small
\centering
\begin{tabular}{llr}
\toprule
\textbf{Dataset} & \textbf{License} & \textbf{Test Ex.} \\
\midrule
Natural Questions & Apache-2.0 & 3,610 \\
TriviaQA & Apache-2.0 & 11,313 \\
WebQuestions & CC BY 4.0 & 2,032 \\
CuratedTREC & - & 694 \\
SQuAD & CC BY-SA 4.0 & 10,570 \\
EntityQs & MIT & 22,075 \\
\bottomrule
\end{tabular}
\caption{The license and number of test example in each of the datasets used in the paper.}
\label{tab:dataset_stats}
\end{table}
Table~\ref{tab:beir} gives the results of our method on the BEIR and MTEB benchmarks for all 19 datasets (complements Table~\ref{tab:zero-shot}). 
Table~\ref{tab:zero-shot-k1}, Table~\ref{tab:zero-shot-k5} and Table~\ref{tab:zero-shot-k100} give the zero-shot results for $k\in\{1,5,100\}$, respectively (complement Table~\ref{tab:zero-shot}).

\section{Dataset Statistics \& Licenses}
Table~\ref{tab:dataset_stats} details the license and number of test example for each of the six open-domain datasets used in our work. 
For the BEIR benchmark, we refer the reader to \citet{beir} for number of examples and license of each of their datasets. 

\section{Computational Resources}

Our method (LE) does not involve training models at all. Our computational resources have been used to evaluate LE on the BEIR benchmark, \textit{i.e.}, computing passage embeddings for each corpus and each model. We used eight Quadro RTX 8000 GPUs. Each experiment took several hours.

\begin{table*}[t]
\centering
\small
\begin{tabular}{lccccc}
\toprule
\textbf{Model} &
\textbf{EntityQs} & \textbf{TriviaQA} & \textbf{~~~~WQ~~~~} & \textbf{~~TREC~~} & \textbf{SQuAD}  \\
\midrule
BM25 & \underline{43.5} & 46.3 & 18.9 & 34.6 & \underline{36.7} \\
BM25 (BERT/MPNet Vocabulary) $\quad\quad$ & 37.6 & 45.4 & 19.2 & 33.0 & 35.6 \\
\midrule
\midrule
DPR  & 24.3 & 37.3 & 30.5 & 51.3 & 16.0 \\
DPR + LE & \textbf{38.3} & \textbf{45.8} & \textbf{35.0} & \textbf{\underline{54.6}} & \textbf{22.8} \\
\midrule
S-MPNet & 22.7 & 42.9 & 30.9 & 51.0 & 25.8 \\
S-MPNet + LE & \textbf{37.3} & \textbf{\underline{47.3}} & \textbf{\underline{37.1}} & \textbf{54.0} & \textbf{30.0} \\
\midrule
Spider & 35.0 & 41.7 & 22.3 & 38.2 & 22.2 \\
Spider + LE & \textbf{40.7} & \textbf{43.7} & \textbf{27.8} & \textbf{43.2} & \textbf{23.5} \\
\bottomrule
\end{tabular}
\caption{Top-1 retrieval accuracy in a ``zero-shot'' setting (i.e., datasets were not used for model training), complementary to Table~\ref{tab:zero-shot}.
LE stands for \textit{lexical enrichment} (our method; \S\ref{sec:token_forget_method}), that enriches query and passage representation with lexical information.
BM25 (BERT Vocabulary) refers to a model that operates over tokens from BERT's vocabulary, rather than words. For each model and dataset, we compare the enriched (LE) model with the original, and mark in bold the better one from the two. We underline the best overall model for each dataset.
}
\label{tab:zero-shot-k1}
\end{table*}


\begin{table*}[t]
\centering
\small
\begin{tabular}{lccccc}
\toprule
\textbf{Model} &
\textbf{EntityQs} & \textbf{TriviaQA} & \textbf{~~~~WQ~~~~} & \textbf{~~TREC~~} & \textbf{SQuAD}  \\
\midrule
BM25 & \underline{61.0} & 66.3 & 41.8 & 64.6 & \underline{57.5} \\
BM25 (BERT/MPNet Vocabulary) $\quad\quad$ & 55.1 & 65.6 & 42.3 & 62.5 & 56.1 \\
\midrule
\midrule
DPR  & 38.1 & 57.0 & 52.7 & 74.1 & 33.4 \\
DPR + LE & \textbf{53.8} & \textbf{64.8} & \textbf{57.7} & \textbf{79.5} & \textbf{42.3} \\
\midrule
S-MPNet & 42.7 & 66.1 & 58.8 & 79.7 & 49.5 \\
S-MPNet + LE & \textbf{56.8} & \textbf{\underline{68.5}} & \textbf{\underline{61.6}} & \textbf{\underline{81.4}} & \textbf{53.2} \\
\midrule
Spider &  54.5 & 63.6 & 46.8 & 65.9 & 43.6 \\
Spider + LE &  \textbf{58.0} & \textbf{64.4} & \textbf{52.2} & \textbf{70.0} & \textbf{44.9} \\
\bottomrule
\end{tabular}
\caption{Top-5 retrieval accuracy in a ``zero-shot'' setting (i.e., datasets were not used for model training), complementary to Table~\ref{tab:zero-shot}.
LE stands for \textit{lexical enrichment} (our method; \S\ref{sec:token_forget_method}), that enriches query and passage representation with lexical information.
BM25 (BERT Vocabulary) refers to a model that operates over tokens from BERT's vocabulary, rather than words. For each model and dataset, we compare the enriched (LE) model with the original, and mark in bold the better one from the two. We underline the best overall model for each dataset.
}
\label{tab:zero-shot-k5}
\end{table*}

\begin{table*}[t]
\centering
\small
\begin{tabular}{lccccc}
\toprule
\textbf{Model} &
\textbf{EntityQs} & \textbf{TriviaQA} & \textbf{~~~~WQ~~~~} & \textbf{~~TREC~~} & \textbf{SQuAD}  \\
\midrule
BM25 & \underline{80.0} & 83.2 & 75.5 & 90.3 & \underline{82.0} \\
BM25 (BERT/MPNet Vocabulary) $\quad\quad$ & 76.6 & 83.0 & 76.0 & 90.5 & 81.1 \\
\midrule
\midrule
DPR  & 63.2 & 78.7 & 78.3 & 92.1 & 65.1 \\
DPR + LE & \textbf{76.1} & \textbf{82.9} & \textbf{82.1} & \textbf{93.5} & \textbf{74.0} \\
\midrule
S-MPNet & 71.7 & 84.8 & 83.0 & \textbf{\underline{95.1}} & 78.4 \\
S-MPNet + LE & \textbf{78.6} & \textbf{\underline{85.1}} & \textbf{\underline{83.8}} & 95.0 & \textbf{80.7} \\
\midrule
Spider & 77.4 & 83.5 & 79.7 & \textbf{92.8} & 76.0  \\
Spider + LE & \textbf{78.9} & \textbf{83.8} & \textbf{81.5} & 92.2 & \textbf{77.8} \\
\bottomrule
\end{tabular}
\caption{Top-100 retrieval accuracy in a ``zero-shot'' setting (i.e., datasets were not used for model training), complementary to Table~\ref{tab:zero-shot}.
LE stands for \textit{lexical enrichment} (our method; \S\ref{sec:token_forget_method}), that enriches query and passage representation with lexical information.
BM25 (BERT Vocabulary) refers to a model that operates over tokens from BERT's vocabulary, rather than words. For each model and dataset, we compare the enriched (LE) model with the original, and mark in bold the better one from the two. We underline the best overall model for each dataset.
}
\label{tab:zero-shot-k100}
\end{table*}
\begin{table*}[t]
\centering
\small
\begin{tabular}{lcccccccc}
\toprule
\multirow{2.5}{80pt}{\textbf{Dataset}}  & \multicolumn{2}{c}{\textbf{DPR}}  && \multicolumn{2}{c}{\textbf{Spider}} && \multicolumn{2}{c}{\textbf{S-MPNet}} \\
\cmidrule(lr){2-3} \cmidrule(lr){5-6} \cmidrule(lr){8-9}
& \textbf{Original} & ~~\textbf{+ LE}~~ && \textbf{Original} & ~~\textbf{+ LE}~~ && \textbf{Original} & ~~\textbf{+ LE}~~ \\
\midrule
MS MARCO & 18.4 & 20.9 && 14.6 & 16.2 && 40.0 & 40.3 \\
TREC-COVID & 22.2 & 30.8 && 30.5 & 32.0 && 51.0 & 51.3 \\
NFCorpus & 15.7 & 19.0 && 27.4 & 26.2 && 33.4 & 33.6 \\
NQ & 51.3 & 49.8 && 12.6 & 17.0 && 52.2 & 52.8 \\
HotpotQA & 32.6 & 37.7 && 40.4 & 43.1 && 45.2 & 48.3 \\
FiQA-2018 & 10.5 & 13.0 && ~~1.0 & 11.2 && 49.3 & 49.8 \\
ArguAna & 10.8 & 14.1 && 31.2 & 31.0 && 39.6 & 49.2 \\
Touché-2020 & 13.1 & 15.8 && ~~4.2 & ~~6.4 && 21.0 & 21.5 \\
CQADupStack & 12.7 & 18.0 && 21.3 & 21.7 && 44.6 & 44.7 \\
Quora & 16.8 & 42.4 && 73.0 & 75.6 && 87.0 & 87.3 \\
DBPedia & 26.9 & 28.5 && 20.0 & 22.3 && 34.1 & 34.8 \\
SCIDOCS & ~~7.4 & 10.1 && 13.1 & 12.8 && 23.6 & 23.5 \\
FEVER & 52.7 & 54.7 && 30.2 & 34.3 && 59.0 & 60.0 \\
Climate-FEVER & 18.2 & 22.9 && 12.4 & 22.4 && 23.1 & 23.6 \\
SciFact & 26.9 & 36.1 && 63.6 & 59.8 && 65.2 & 65.3 \\
BioASQ & 11.6 & 17.6 && 21.0 & 22.3 && 21.5 & 22.3 \\
Signal-1M (RT) & 13.6 & 21.1 && 25.3 & 26.1 && 24.9 & 25.3 \\
TREC-NEWS & 19.1 & 21.3 && 29.3 & 31.3 && 50.7 & 50.7 \\
Robust04 & 22.4 & 22.7 && 36.4 & 35.9 && 50.0 & 50.0 \\
\midrule
Avg. (MTEB: Retrieval) & 22.4 & \textbf{27.6} && 26.4 & \textbf{28.8} && 44.6 & \textbf{45.7} \\
Avg. (BEIR) & 21.4 & \textbf{26.4} && 27.4 & \textbf{29.5} && 43.1 & \textbf{44.1} \\
\bottomrule
\end{tabular}
\caption{Retrieval results measured by nDCG@10 on BEIR (all datasets except MS MARCO) and the retrieval cluster of MTEB (first 15 datasets).
LE stands for \textit{lexical enrichment} (our method; \S\ref{sec:token_forget_method}), that enriches query and passage representation with lexical information.
}
\label{tab:beir}
\end{table*}


\end{document}